\documentclass{article}

\usepackage{ulem}

\usepackage{arxiv}
\usepackage{float}
\usepackage[utf8]{inputenc} 
\usepackage[T1]{fontenc}    
\usepackage{amsfonts}       
\usepackage{doi}
\usepackage{amssymb}
\usepackage{amsthm}
\usepackage{graphicx}                           
\usepackage{dcolumn}  
\usepackage[version=3]{mhchem}  
\usepackage[T1]{fontenc}                    

\usepackage{epsfig}
\usepackage{epstopdf}
\usepackage{amsbsy}
\usepackage{multirow}
\usepackage{bm}
\usepackage{siunitx}
\usepackage {CJK}
\usepackage{textcomp}
\usepackage{comment}
\usepackage{nomencl}
\usepackage{parskip}   
\usepackage{tabularx}
\usepackage{caption} 
\usepackage{todonotes}
\makeatletter                                   
\renewcommand{\fnum@figure}{Figure \thefigure} 
\makeatother                                    
\usepackage[ruled,vlined]{algorithm2e}
\usepackage{mdframed}

\usepackage{bbm}

\title{Lagrangian Neural Networks for Reversible Dissipative Evolution
}

\author{ {Veera Sundararaghavan}\thanks{Corresponding author: Prof. Sundararaghavan, Email: veeras@umich.edu, Tel: 734-615-7242} \\
	Department of Aerospace Engineering\\
	University of Michigan\\
	Ann Arbor, MI \\
	\texttt{veeras@umich.edu} \\
	\And
	{\hspace{1mm}Megna N Shah and Jeff P Simmons} \\
	Materials and Manufacturing Directorate\\
	Air Force Research Laboratory\\
	Wright Patterson Air Force Base, OH \\
	\texttt{megna.shah.1@us.af.mil} \\
	\texttt{jeff.simmons.3@afrl.af.mil} \\
}

\begin{document}
\maketitle

\begin{abstract}
There is a growing attention given to utilizing Lagrangian and Hamiltonian mechanics with network training in order to incorporate physics into the network. Most commonly, conservative systems are modeled, in which there are no frictional losses, so the system may be run forward and backward in time without requiring regularization. This work addresses systems in which the reverse direction is ill-posed because of the dissipation that occurs in forward evolution. The novelty is the use of  Morse--Feshbach Lagrangian, which models dissipative dynamics by doubling the number of dimensions of the system in order to create a ‘mirror’ latent representation that would counterbalance the dissipation of the observable system, making it a conservative system, albeit embedded in a larger space. We start with their formal approach by redefining a new Dissipative Lagrangian, such that the unknown matrices in the Euler-Lagrange's equations arise as partial derivatives of the Lagrangian with respect to only the observables. We then train a network from simulated training data for dissipative systems such as Fickian diffusion that arise in materials sciences. It is shown by experiments that the systems can be evolved in both forward and reverse directions without regularization beyond that provided by the Morse- Feshbach Lagrangian. Experiments of dissipative systems, such as Fickian diffusion, demonstrate the degree to which dynamics can be reversed.

\end{abstract}

\section{Introduction}
\label{sec:Introduction}

Physics of mechanical systems can be reduced to a Lagrangian formulation, where optimal paths taken between points A and B are determined as the function that makes the action integral stationary. The foundation of this approach is Noether's theorem\cite{noether1983invariante}, which connects the stationary modes of the action integral to conserved quantities. The variational problem leading to the optimal path utilizes a search over a space of `admissible functions,' typically continuous or continuous with continuous derivatives~\cite{gelfand2000calculus}. This space has, within it, manifolds possessing symmetries. Symmetries lead to invariances or, as they are known in physics, conserved quantities. Thus, the solution to the Euler-Lagrange equation exists where the conserved quantities coincide.

While the Lagrangian formulation of physics has been very successful for describing systems in which there are conserved quantities,  real-world systems often contain dissipation due to frictional effects. In a classic work, Morse and Feshbach\cite{morse1954methods} described a purely formal approach to using Lagrangian methods for describing dissipative systems, that is, systems in which the quantity of interest is \emph{not} conserved. In their approach, they constructed a space in which the number of coordinates was doubled with the observable half of these coordinates being what they are in the real physical world and the latent half of coordinates assuming whatever value necessary to make the whole system obey a conservation law. Thus, it would obey Noether's theorem with this dimension-doubled space and enjoy all of the advantages that conservative systems do. Lagrangian mechanics has attracted a growing interests for describing conservative systems via machine learning as of late~\cite{cranmer2020lagrangian,xiao2024generalized,lutter2019deep}. We undertook this study in order to evaluate the effectiveness of the Morse and Feshbach approach for dissipative systems under machine learning. Such systems include frictional or diffusional dissipation, the latter being systems invoving thermal, viscous, or chemical diffusion via Fick's law.

Before proceeding, it is important to point out that the problem is timely, enabling optimization or autonomous control of processing conditions in chemical or materials science applications\cite{nikolaev2016autonomy}. It is well appreciated that the popular diffusion models\cite{sohl2015deep} used in generative modeling achieve, in some sense, our goals here. But, for optimization and iterative approaches, it is necessary for minor adjustments to be very computationally fast: it is not simply enough to be able to reverse a dissipative process-it is more desirable to have a representation of these processes in which equal facility is achieved in both the forward and reverse time directions. It should also be appreciated that, while this study will use diffusion, we employ \emph{Fick's law} coarse-grained diffusion\cite{fick1855ueber,fourier1888theorie}, whereas diffusion models use \emph{Brownian} diffusion\cite{gillespie2013simple}. While these can be related\cite{lebowitz1982microscopic}, Brownian diffusion is somewhat tangential to our purpose and it will not be pursued in this study.

\subsection{Related Work}

Cranmer, \textit{et al.} \cite{cranmer2020lagrangian} and Lutter et al. \cite{lutter2019deep} developed a generalized way of fitting a Lagrangian to a dataset and applied this to systems where quantities (e.g. energy) were conserved. Similar techniques exist in the context of learning Hamiltonians~\cite{greydanus2019hamiltonian,toth2019hamiltonian}. However, these techniques are only for conserved systems without energy loss. Recently there has been significant interest in modeling real world dissipative systems using this concept. Sosanya and Greydanus developed ‘dissipative hamiltonian neural networks’ \cite{sosanya2022dissipative} which describes dissipative systems using a Hamiltonian and a dissipative term. A similar method for Lagrangian neural networks~\cite{xiao2024generalized} has been recently developed. However, these methods are specific to modeling the specific case of Rayleigh dissipation in the equations of motion and do not have wide applicability to other physics such as diffusion. 

For dissipative systems, a more general theory builds on the approach proposed by Morse and Feshbach\cite{morse1954methods} for treating dissipative systems with a Lagrangian approach. They start with the differential equation for an oscillator with friction and propose a `purely formal expression' for the Lagrangian. The key idea is to double the number of coordinates~\cite{bateman1931dissipative} so that, if energy is lost through dissipation, it is transferred to a `mirror image' (latent) space. In this way, the energy of the doubled dimension system remains constant. 

While conceptually correct in that to model dissipation, the loss of energy to environment needs to be considered, Morse and Feshbach relied on a ingenious mathematical trick that ensures the correct form of equations is achieved in the Euler--Lagrange's equations. While there are other approaches employing coordinate-doubling schemes\cite{al2020hamiltonian,anthony1989unification,kotowski1992hamilton,muschik2000variational}, it appears that the theory behind coordinate doubling is still in development at this point~\cite{szegleti2020dissipation,bersani2021lagrangian} and we simply use Morse and Feshbach as a starting point. We also extend this idea, in the case of diffusion, so that the coordinate doubled system represents a quantity that is invariant and includes the diffusion problem among the observable coordinates.

\subsection{Contribution}

The contribution of this paper is to extend Lagrangian mechanics to include dissipative systems so that evolution of these systems may be represented in a phase space framework, allowing the state of the system at any point in time to be reconstructed from its phase space trajectory. While many approaches for regularized time reversal of these types of systems exist, to our knowledge, this is the only machine learning approach that is applicable to general dissipative systems such as diffusion.

Our new contributions include: (1) increasing the generality of Morse and Feshbach's formal Lagrangian expression, (2) Proposing a new `Dissipative Lagrangian'($\mathcal{D}$) that can capture  evolution purely in terms of observables, and (3) using a network to learn the form of the dissipative Lagrangian in the way of Cranmer, \textit{et al.}\cite{cranmer2020lagrangian}

All codes will be made freely available at publication time.

\section{Preliminary Concepts}

Morse and Feshbach started with the dissipative harmonic oscillator, governed by:
\begin{equation}
    M\ddot{x} + C\dot{x} + Kx = 0
    \label{eq:GeneralHarmonicOscillator}
\end{equation}
where $M\in \Re$ is a `mass,' allowing for inertia of a system, $C\in \Re$ is a damping factor, $K\in\Re$ is a `stiffness,' and $x\in\Re$ is a generalized displacement variable representing the change from an equilibrium value.

If $C\neq 0$, the system is not conservative and conventional Lagrangian would fail. To address this, the authors proposed a `purely formal expression:'
\begin{equation}
    \mathcal{L} = M(\dot{x}\dot{\eta}) - \frac{1}{2}C(\eta\dot{x} - x\dot{\eta}) - Kx\eta
    \label{Eq:MnF-purelyformal}
\end{equation}
where $\mathcal{L}$ is the Morse--Feshbach Lagrangian and $\eta\in\Re$ defines displacements of a `mirror-image oscillator' with negative friction.
In the usual fashion, they construct the momenta as derivatives of the Lagrangian, from $p_x \equiv \frac{\partial \mathcal{L}}{\partial \dot{x}}$ and $p_{\eta} \equiv \frac{\partial \mathcal{L}}{\partial \dot{\eta}}$:
\begin{align}
    p_x = M\dot{\eta} - \frac{1}{2}C\eta && p_\eta = M\dot{x} - \frac{1}{2}Cx
\end{align}
where $p_x$ and $p_\eta$ are the momenta in the observable and mirror-image spaces, respectively. Note that the observable momentum is a function of the mirror-image variables and the mirror-image momentum is a function of the observable coordinates. This is, as they pointed out, an unusual situation, but, the approach was justified by the fact that it could reproduce the original equations of motion via the pair of Euler-Lagrange equations:
\begin{align}
\frac{d}{dt} \frac{\partial \mathcal{L}}{\partial \dot{\eta}} - \frac{\partial \mathcal{L}}{\partial \eta} = 0 &&  \frac{d}{dt} \frac{\partial \mathcal{L}}{\partial \dot{x}} - \frac{\partial \mathcal{L}}{\partial x} = 0 
 \label{Eq:mnfel}
\end{align}
Substituting in the Lagrangian in Equation \ref{Eq:MnF-purelyformal}, the first equation provides equation of motion for the real system (Equation \ref{eq:GeneralHarmonicOscillator}) and the second evolves the mirror system with negative damping. Hence, we have:
\begin{align}
    M\ddot{x} + C\dot{x} + Kx = 0   &&  M\ddot{\eta} - C\dot{\eta} + K\eta = 0
\end{align}
as the governing equations in the observable and mirror-image space, respectively.

\section{Learning Lagrangians for Dissipative Problems}

\subsection{Lagrangian Formulation of Dissipative Mechanics}

We extend Morse and Feshbach's formal expression (Equation (\ref{Eq:MnF-purelyformal}) to a multidimensional system so that 
an image representing a field (composition or temperature) containing $N$ pixels, has $N$ dimensions in the observable space. By replacing the scalar multiplied products with bilinear forms of matrices in this equation, we have:
\begin{equation}
\mathcal{L} = \sum_{ij}\left[\dot{\eta_i} M_{ij} \dot{x_j} -\frac{1}{2} (\dot{x_i}C_{ij}\eta_j - \dot{\eta}_{i}C_{ij}x_j) - \eta_i K_{ij} x_j\right]
\label{eq:M&FModel}
\end{equation}
where we have redefined $M_{ij},C_{ij},K_{ij}\in \Re^{N\times N}$, we have also redefined $\eta_i,x_i\in \Re^N$, where $i$ is an index over the dimensions of the multi-dimensional space.

The Euler Lagrange equations (Eq. \ref{Eq:mnfel}) for the proposed Lagrangian reduces to:
\begin{align}
    \sum_{j}\left[M_{ij}\ddot{x_j} + \frac{1}{2}(C_{ij} + C_{ji})\dot{x_j} + K_{ij}x_j\right] = 0    && \sum_{j}\left[M_{ji}\ddot{\eta_j} - \frac{1}{2}(C_{ij}+C_{ji})\dot{\eta_j} + K_{ji}\eta_j\right] = 0 \nonumber
\label{eq:motion}
\end{align}

Note that all occurrances of $C$ in the Euler--Lagrange equations are in the form of its symmetric part, $C_{sym} = \frac{1}{2} (C_{ij}+C_{ji})$, so that there $C$ need not have any special structure usually applied in conventional cases.
We prefer to admit to only using observables for training, as was done by Sosanya, \textit{et al.}\cite{sosanya2022dissipative} in the case of dissipative Hamiltonian networks, so we introduce a new  dissipative Lagrangian ($\mathcal{D}$) that makes this possible:
\begin{equation}
\mathcal{D} = \sum_{ij}\left[ \frac{1}{2}\dot{x}_iM_{ij}\dot{x}_j + \frac{1}{2}\dot{x}_iC_{ij}x_j  + \frac{1}{2}x_iK_{ij}x_j\right]
\label{Mdiff2firstcut-mod}
\end{equation}

Rewriting the Morse-Feshbach Lagrangian in Eq. \ref{eq:M&FModel} allows us to eliminate the explicit dependence on the parameters $M$, $C$, and $K$ through the following steps. First:
\begin{equation}
\mathcal{L} = \sum_{ij}\left[\dot{\eta_i} (M_{ij} \dot{x_j}+\frac{1}{2}C_{ij}x_j) - \eta_i (K_{ij} x_j +\frac{1}{2} C_{ji}\dot{x_j}) \right]
\label{eqsplit}
\end{equation}
Then, note that the terms in parentheses in this equation can be directly related to simple derivatives of $\mathcal{D}$. Substituting:
\begin{equation}
\mathcal{L} = \sum_i\left[\dot{\eta_i} \frac{\partial \mathcal{D}}{\partial \dot{x_i}} - \eta_i \frac{\partial \mathcal{D}}{\partial x_i}\right]
\label{eqsplit2n}
\end{equation}
Thus, the following equalities are satisfied:\\
\begin{equation}
\frac{\partial \mathcal{L}}{\partial \eta_i} = -\frac{\partial \mathcal{D}}{\partial x_i}, \quad \frac{\partial \mathcal{L}}{\partial \dot{\eta}_i} =\frac{\partial \mathcal{D}}{\partial \dot{x}_i}
\label{Mdiff2firstcutn}
\end{equation}

Substituting Equation (\ref{Mdiff2firstcutn}) into the Euler-Lagrange equations, we obtain a vector form of Equation (\ref{Eq:mnfel}):
\begin{align}
\frac{d}{dt} \frac{\partial \mathcal{D}}{\partial \dot{x}_i} + \frac{\partial \mathcal{D}}{\partial x_i} = 0  &&  \frac{d}{dt} \frac{\partial \mathcal{L}}{\partial \dot{x}_i} - \frac{\partial \mathcal{L}}{\partial x_i} = 0 
\label{eln}
\end{align}
Unlike the original equation proposed by Morse and Feshbach in Equation \ref{Eq:mnfel}, the first equation that models the dynamics of the real system is now only related to the observables in the real system. However, the linear differential operator of the first Euler-Lagrange equation now relates to the adjoint of that in the second Euler-Lagrange equation. For problems with constant coefficients, the matrices $M$, $K$ and the symmetric part of $C$ can be retrieved from Equation~\ref{Mdiff2firstcut-mod} as:
\begin{align}
M_{ij} = \frac{\partial^2 \mathcal{D}}{\partial \dot{x_i}\partial\dot{x_j}} && K_{ij} = \frac{\partial^2 \mathcal{D}}{\partial x_i\partial x_j} && C_{sym} = \frac{1}{2}\left(C_{ij} + C_{ji}\right) = \frac{\partial^2 \mathcal{D}}{\partial \dot{x_i}\partial x_j } + \frac{\partial^2 \mathcal{D}}{ \partial x_i\partial \dot{x_j}}
\end{align}

\subsection{Modeling Diffusive Systems}

While the above formulation is for classical mechanics, the approach is general enough to model other equations such as diffusion. In this case, we replace the generalized coordinate $x$ with $c$, the concentration. The Fickian diffusion equation, after appropriate numerical discretization  such as finite differencing, can be written in a matrix form as:
\begin{equation}
    \dot{c_i} + \sum_j K_{ij}c_j = 0
    \label{eq:diffmat}
\end{equation}
where $K\in \Re^{N\times N}$ is a stiffness matrix, that contains the diffusion coefficients, $c\in\Re^N$ is the vector of unknown concentrations and $\dot{c}\in\Re^N$ are its time derivatives. The Morse-Feshbach Lagrangian is given by: 
\begin{equation}
\mathcal{L} = \sum_{i}\frac{1}{2}(\dot{\eta_i}c_i - \eta_i\dot{c_i})- \sum_{ij}\eta_i K_{ij} c_j  
\label{eqsplitd}
\end{equation}
In the context of a diffusing system, the auxiliary variable, $\eta$, can be understood as an `undiffuser', so that the two systems, together, are conservative. While the observable, $c$, dissipates, $\eta$ `un-dissipates,' allowing complete recovery of the state of the system at every moment. 
As above, we define a dissipative Lagrangian:

\begin{equation}
\mathcal{D} = \sum_{i}\frac{1}{2}\dot{c}_ic_i  + \sum_{ij}\frac{1}{2}c_iK_{ij}c_j
\label{Mdiff}
\end{equation}

The same sequence of steps as in Equations (\ref{eqsplit2n})-(\ref{eln}) can be followed to arrive at the Euler--Lagrange equations which further reduces to the evolution equations for concentration and the auxiliary variable:
\begin{align}
    \dot{c}_i + \sum_j K_{ij}c_j  = 0   &&  \dot{\eta}_i - \sum_j  K_{ij}\eta_j  = 0
\end{align}

For equations with constant coefficients, the matrix $K$ follows directly from Equation (\ref{Mdiff}):
\begin{align}
K_{ij} = \frac{\partial^2 \mathcal{D}}{\partial c_i\partial c_j} \label{kdiff}
\end{align}

\section{Neural network architecture}

The neural network algorithm was implemented in MATLAB using the dlgradient function that allows higher order derivatives of the output of the neural network with respect to the inputs.The network consists of two hidden layers of 200 neurons each and a final output layer of 1 neuron. A fully connected network was used. The network was trained using the Adam optimizer for 1000 epochs per batch. A batch size of 1000 was used for all datasets. 

\begin{figure}[H]
    \centering
    \includegraphics[width=0.8\linewidth]{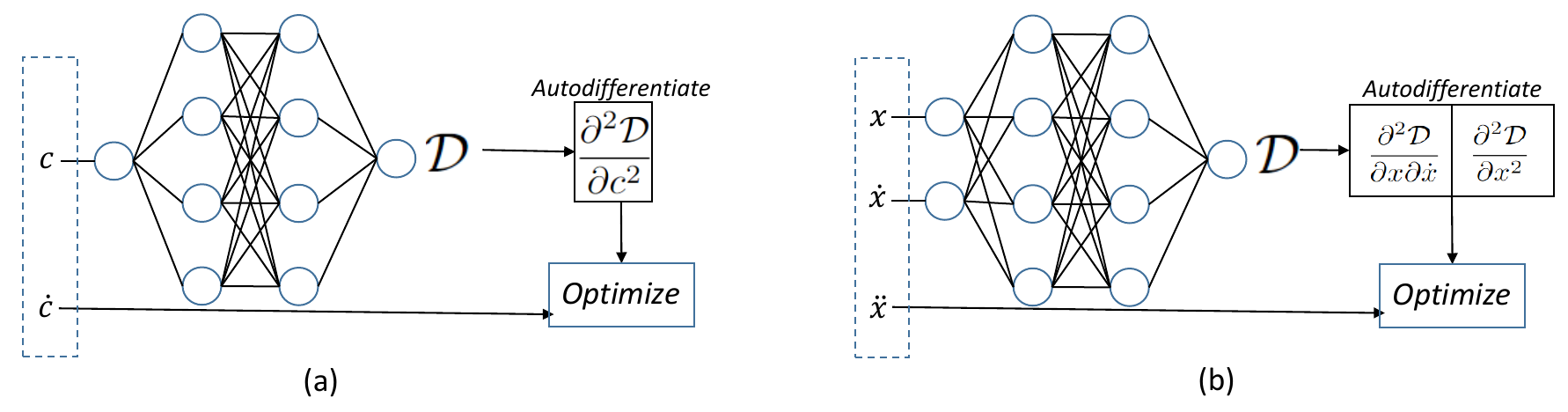}
    \caption{Neural network architectures for  (a) diffusion problem and (b) the dissipative mechanics problems}
    \label{netfig}
\end{figure}

In the dissipative Lagrangian Neural Network (DLNN) implementation of the diffusion equation (see Figure ~\ref{netfig}a), the unknown constant coefficient matrix $K$ represented using the dissipative Lagrangian $\mathcal{D}$ using Eq. \ref{kdiff}. The loss function is directly based on the matrix equation  \ref{eq:diffmat} , using the training data ($c_{data},\dot{c}_{data}$) for the observables.
\begin{equation}
L_{DLNN} = ||\dot{c}_{data,i} + \sum_j \frac{\partial^2 \mathcal{D}}{\partial c_i\partial c_j} c_{data,j}||
\end{equation}

The network takes as input vector $c_{data}$, which are known concentration values ($N\times M$ pixel values of an image) at time t. Each training set contains doublet ($c_{data}, \dot{c}_{data}$) at a given time step. The rate $\dot{c}_{data}$ is used in training when calculating the loss function but is not part of the network inputs. The output is a scalar, the dissipative Lagrangian.

Once the network is trained, new evolution trajectories can be simulated by using a new initial condition, followed by calculating the rate $\dot{c}$ using autodifferentiation of predicted $\mathcal{D}$ and the data is used to march the solution $c$ in time using ODE solvers. The new $c$ is then passed again to the network and the time stepping continues using the new $\dot{c}$ found using autodifferentiation. 

For the dissipative mechanics problems, two initial condition vectors (position and velocity) are needed. In this case, the input layer takes in both $x$ and $\dot{x}$ at time t as shown in Figure \ref{netfig}(b), the network thus contains twice the number of neurons as the number of positional degrees of freedom. Each training set contains three vectors ($x, \dot{x}, \ddot{x}$) at a given time step. The acceleration data is only used in the loss function but is not an input to the network. The mass matrix is generally a known quantity and is directly used in the neural network. The unknown stiffness and damping matrices are obtained via autodifferentiation of the output $\mathcal{D}$ and used in the following loss function:
\begin{equation}
L_{DLNN} = ||\sum_j M_{ij}\ddot{x}_{data,j} + \sum_{j}\left(\frac{\partial^2 \mathcal{D}}{\partial \dot{x_i} \partial x_j } + \frac{\partial^2 \mathcal{D}}{ \partial x_i\partial \dot{x_j}}\right)\dot{x}_{data,j} + \sum_{j}\frac{\partial^2 \mathcal{D}}{\partial x_i\partial x_j} x_{data,j}||
\end{equation}

\section{Experiments}

The goal of this project is to demonstrate an invertible model of dissipative systems. To do this, we first model a mechanical system of only four degrees of freedom. Then, we proceed to the more complex situation of diffusion fields, where the number of degrees of freedom is the number of pixels in an image.

\subsection{Mechanics}

\begin{figure}[H]
    \centering
    \includegraphics[width=0.9\linewidth]{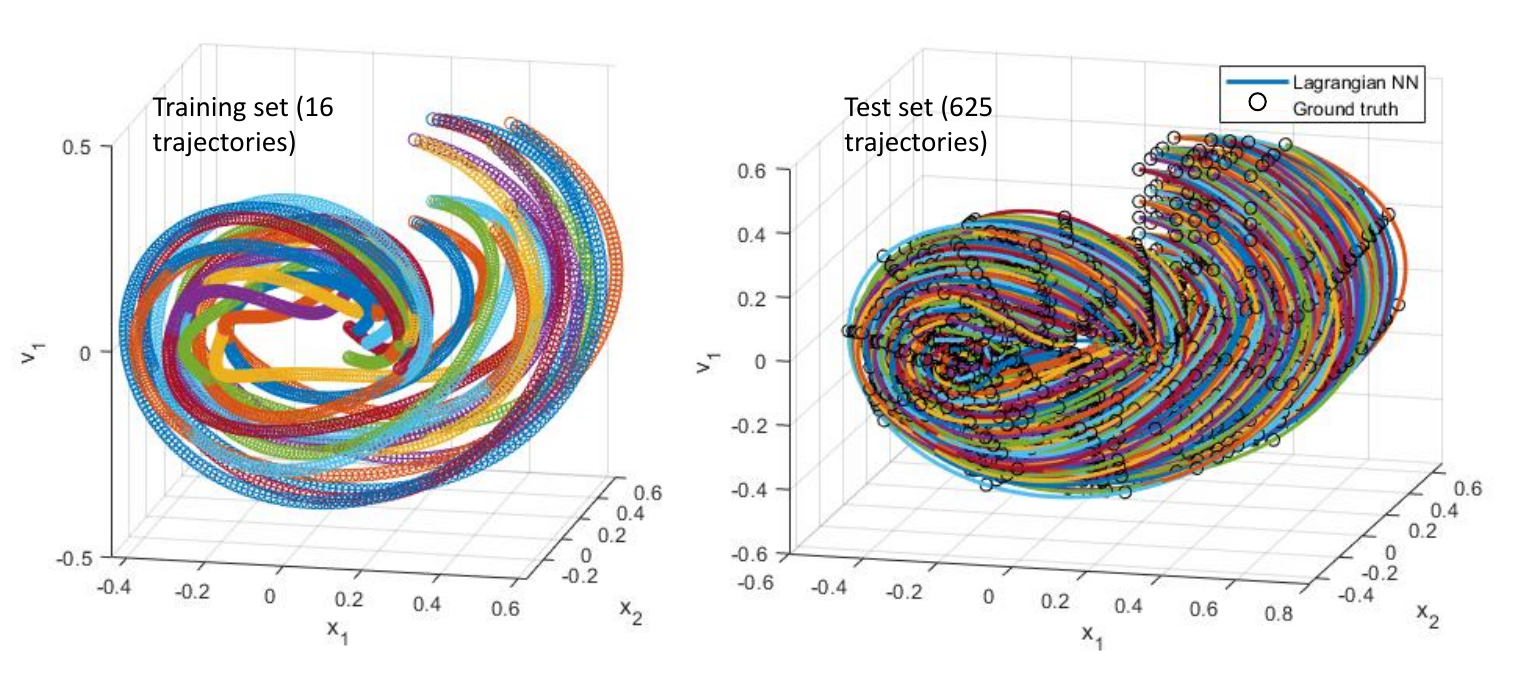}
    \caption{
    4-D phase space of the mass--spring--damper system, projected onto the the 3-D space of $(x_1,x_2,\dot{x}_1)$.
    (a) data used for training containing 16 trajectories (b) DLNN prediction for 625 new trajectories. }
    \label{spring}
\end{figure}

We consider a simple problem of a  mass--spring--damper system with parameters given in Table \ref{tab:MechanicalSystemParameters}. 
Training data contained simulated trajectories, based on time integration using an explicit Runge-Kutta (4,5) formula, the Dormand-Prince pair \cite{dormand1980family}, with initial positions and velocities taken from a square grid formed by all permutations of $x_i,\dot{x}_i\in [0.2,0.4], i\in[1,2]$.
The test data contained 625 new extrapolatory trajectories $x_i, \dot{x}_i\in [0.15,    0.25,0.3,0.35,0.45]$.
Results are shown in Figure~\ref{spring}(a). Figure \ref{spring}(b) shows the test case trajectories predicted by the DLNN and the analytical solution for these trajectories are indicated. The method is able to predict the physical behavior of the system accurately from limited training data. 
\begin{table}
\parbox{.35\linewidth}{
\centering
\begin{tabular}{ccc}
\hline
\textbf{Parameter}   &   \textbf{Value}\\
    \hline
      $K_{11}$   &  1\\
      $K_{22}$    &   1\\
      $K_{12}$    &   -0.4\\
      \hline
      $C_{11}$    &   0.1\\
      $C_{12}$    &   0.1\\
      $C_{22}$    &   0.2\\
      \hline
      $M$ &   $\mathbbm{1}_{2\times 2}$\\
\hline
\end{tabular}
\caption{Parameters used in the simple mass--spring--damper system.}
\label{tab:MechanicalSystemParameters}
}
\hfill
\parbox{.65\linewidth}{
\centering
\begin{tabular}{ccccc}
\hline
Problem&This work& HNN \cite{sosanya2022dissipative}& LNN \cite{xiao2024generalized} & BaselineNN\\
Mechanics&1E-3& 1E-3 & 1E-3 & 7E-3  \\
Diffusion&1E-3& NA& NA & 8E-3\\
\hline
\end{tabular}
\caption{Maximum error comparison of the present work with current techniques for extrapolatory cases. Mechanics problem is a single mass--spring--damper (M=K=C=1) and diffusion is the model in Fig. \ref{baseline}. Refs \cite{sosanya2022dissipative,xiao2024generalized} are not applicable to diffusion equations.}
}
\end{table}
In a conserved system, each trajectory in the phase space follows a unique curve that has a zero curl.  The presence of dissipation, however, adds sources and sinks in the phase space. Prior work in the modeling of dissipative mechanics using neural networks called `dissipative hamiltonian neural networks' (DHNN)\cite{sosanya2022dissipative} employs two variables to describe such situations. The Hamiltonian is used to capture the curl--free field in the phase space and a dissipative term is used to capture the sources and sinks in the phase space. 
However, in these examples with constant coefficients, we show that the state can be reduced to one variable $\mathcal{D}$. The notion that a single variable is able to explain the dynamics of the dissipative system can be understood in the context of intrinsic dimensionality (ID) of non--intersecting spiral trajectories, which is  \emph{one}. 

In Table 2, we compare the maximum error in extrapolatory mode for the current model against the model implementations for \cite{sosanya2022dissipative,xiao2024generalized} and a baseline neural network that only predicts the acceleration. The present model is comparable to the published models and is superior to the baseline NN. In the case of diffusion (model later explained in section 5.3), however, these published models are not applicable, and the present model is able to exceed the performance of the baseline NN.

\subsection{Fickian Diffusion}
\subsubsection{Two Pixel}
\begin{figure}[H]
    \centering
    \includegraphics[width=1.0\linewidth]{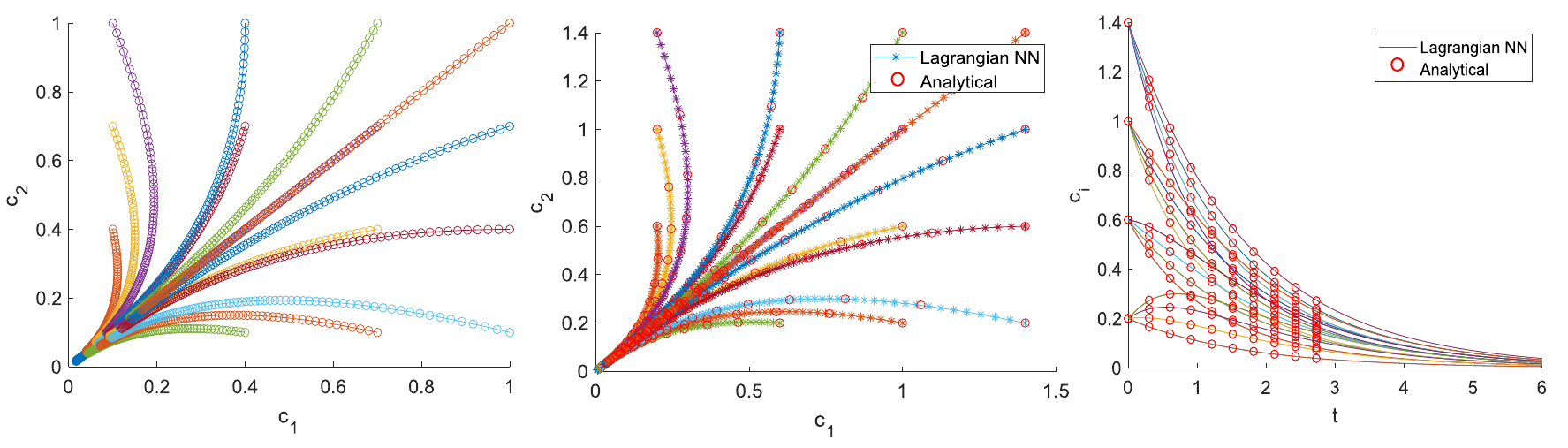}
    \caption{(a) Training data containing 16 trajectories in phase space (b) Predictions of DLNN for the test set in phase space and  (c) concentration versus time upto t =6 s.}
    \label{fig:2px}
\end{figure}

The toy problem for diffusion uses a two pixel model (two concentrations $c_1$ and $c_2$) governed by the same $2 \times 2$ stiffness matrix as in Table. 1. The data was trained using 16 different initial conditions $c_1,c_2 = 0.1$ to $c_1,c_2=1$ in the increments of 0.3 and all combinations of the two, with data upto a time of t=3 seconds. A batch size of 750 was used for the NN training. The phase space of $c_1$ versus $c_2$ is shown for these trajectories in figure \ref{fig:2px}a. The trained DLNN was used to predict 16 new trajectories not in the training set (Figure \ref{fig:2px}b) as well as the predictions were made up to a larger time of t=6 seconds (Figure \ref{fig:2px}c) and the results are compared with solutions from a ODE solver. The results show that the DLNN was able to sufficiently generalize the data to new trajectories not in the training set.
\subsection{Comparison to baseline model}
\begin{figure}[H]
    \centering    \includegraphics[width=0.8\linewidth]{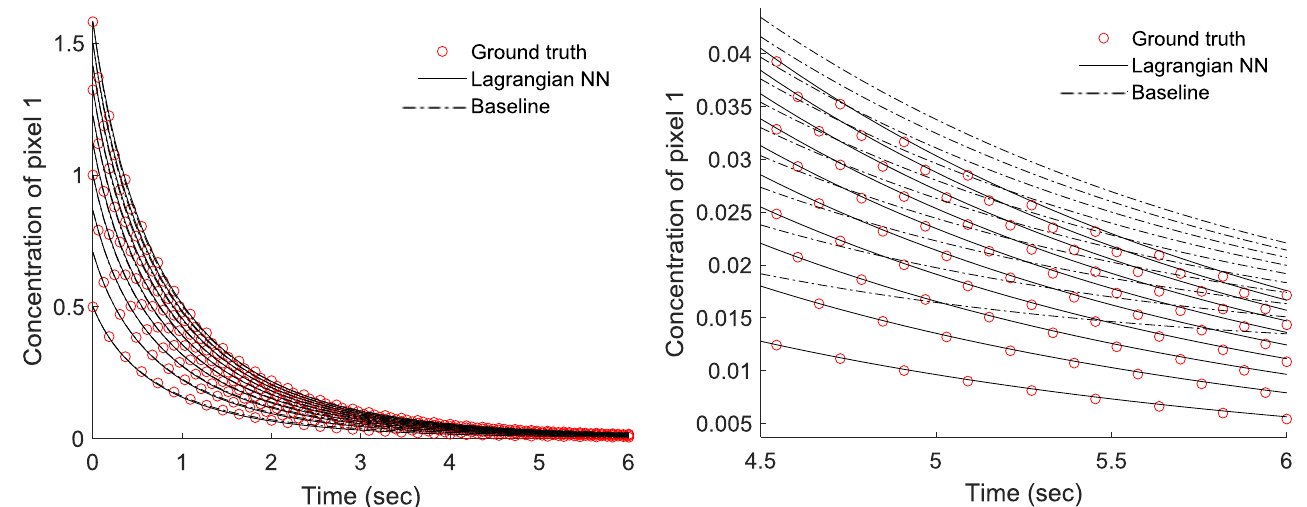}
    \caption{(a) Results from the baseline and DLNN model are compared for an extrapolatory case. (b) The differences between the DLNN and baseline model zoomed in beyond $t=4.5$ seconds}
    \label{baseline}
\end{figure}

The DLNN model is compared to a baseline NN model here. The network for the baseline model has a larger number of output neurons, equal to number of pixels in the model, compared to DLNN that just has one output neuron. The baseline model is trained to minimize the L2 loss between the output and the vector containing known rates of concentration ($\dot{c}_{data}$). For this example, a ten-pixel model was used with the training data corresponding to a case with $K$ being a $10\times10$ Lehmer matrix. Ten different cases ($I = 1,..,10$) were generated, with each case corresponding to an initial concentration vector of $\sqrt I \times [1,1/2,1/3,...,1/10]$. Both the baseline model and the DLNN model were trained to the same L2 loss error of $10^{-5}$, upto a simulation time of 6 seconds. Since the losses measure the deviation from $\dot{c}_{data}$ in both models, this allows a fair comparison of these two models. In particular, we are interested in extrapolatory performance when the initial conditions are halved compared to the training set. The results from the baseline and DLNN model are compared to ground truth in Figure \ref{baseline}(a). 

While the baseline model is able to predict the overall trends, numerically the results from DLNN is significantly superior to the baseline model at larger times. The differences between the DLNN and baseline model are shown alongside the ODE solution at t>4.5 seconds in  Figure \ref{baseline}(b).

\subsection{Sequence of diffusion images}

\begin{figure}[H]
    \centering    \includegraphics[width=1.0\linewidth]{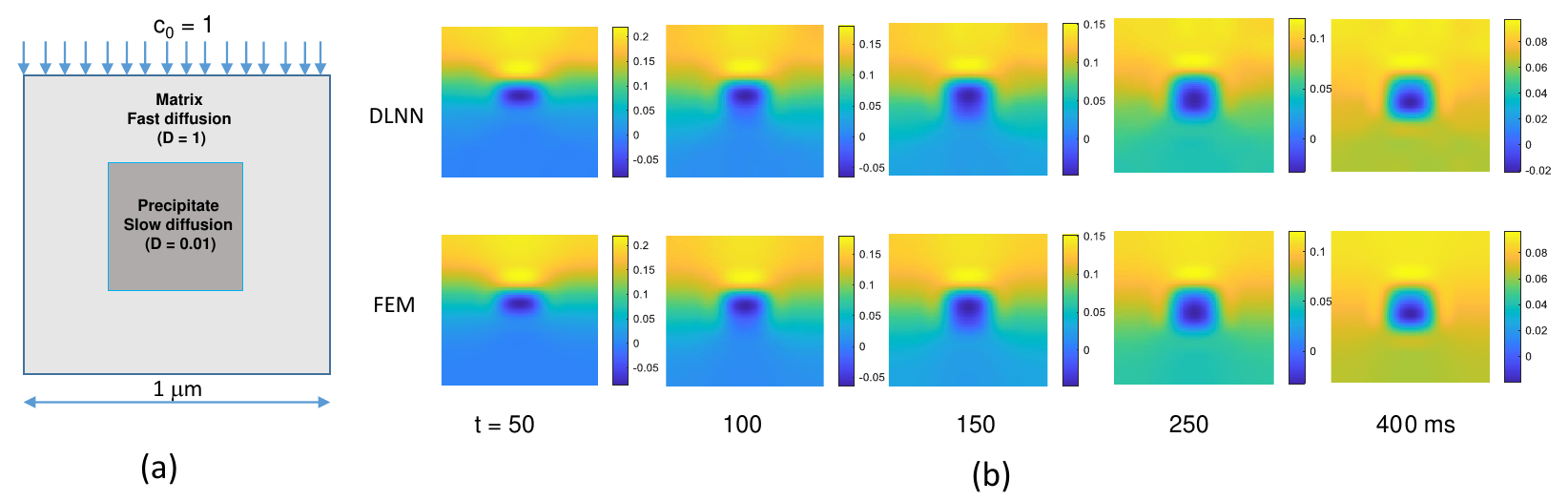}
    \caption{(a) Microstructure model with simulation parameters (b) (top) DLNN simulation at different time steps showing concentration profiles compared to FEM results (bottom)}
    \label{dlnnmicro}
\end{figure}

This example shows modeling of diffusion trajectories in a two phase planar microstructure. The matrix phase is made of a material with high diffusivity (D = 1 $\mu$m$^2$-s) while the precipitate phase is made of a material with low diffusivity (D = 0.01 $\mu$m$^2$-s) as shown in Figure \ref{dlnnmicro}(a). A concentration of 1 mol/$\mu$m$^3$ is applied to the top surface. Zero flux boundary conditions are applied on all other boundaries. The training data is obtained from a finite element model with a 7-by-7 grid measuring 1 $\mu$m by 1 $\mu$m up to a simulation time of 0.5 seconds using 100 time steps. The resulting data in the form of a image sequence containing concentrations at different times and the rate of change in concentration across consecutive frames are recorded in a matrix for use in training. 

\begin{figure}[H]
    \centering    \includegraphics[width=1.0\linewidth]{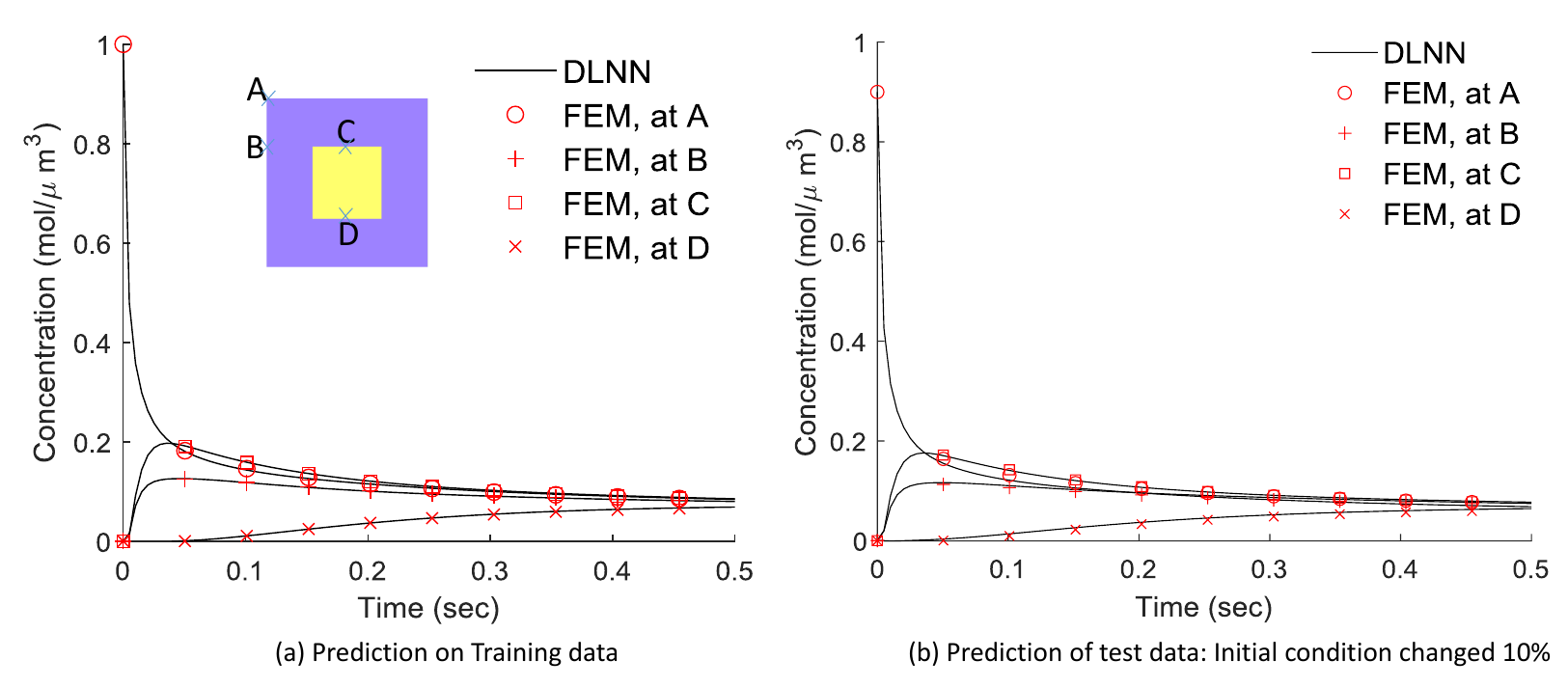}
    \caption{(a) The concentration profiles with respect to time predicted by DLNN are compared to FEM results at four points in the microstructure shown in inset (b) Similar results for an extrapolatory case with a different initial condition.}
    \label{dlnnmicro2}
\end{figure}

The neural network consists of an input layer of 64 neurons and a single hidden layer of 600 neurons. The general rule of thumb followed is to use a hidden layer that is 8 to 10 times larger than the input layer for good training performance. An RMS error of $10^{-5}$ was achieved during training with a batch size of 100. The data was then tested over the training data and the results are shown alongside the actual training images in Figure \ref{dlnnmicro}(b). 
The concentration profiles with respect to time at four points in the microstructure are shown in Figure \ref{dlnnmicro2}(a). The concentration rapidly drops in node A (the boundary where the concentration is initialized) as diffusion takes away the concentration towards the matrix. At all other points the concentration increases. 

To demonstrate an extrapolatory case, the initial concentration at each node was changed by 10 \% to 0.9 mol/$\mu$m$^3$. This implies a 11.1\% RMS change in the initial point in 64D space. The predicted results at the same four points A--D are shown in Figure \ref{dlnnmicro2}(b) alongwith the FEM results for the new initial concentration profile. The overall RMS error in the trajectory in the 64D space was 3.7\%.

\subsection{Inverse diffusion following Lagrangian trajectory}

\begin{figure}[H]
    \centering
    \includegraphics[width=1.0\linewidth]{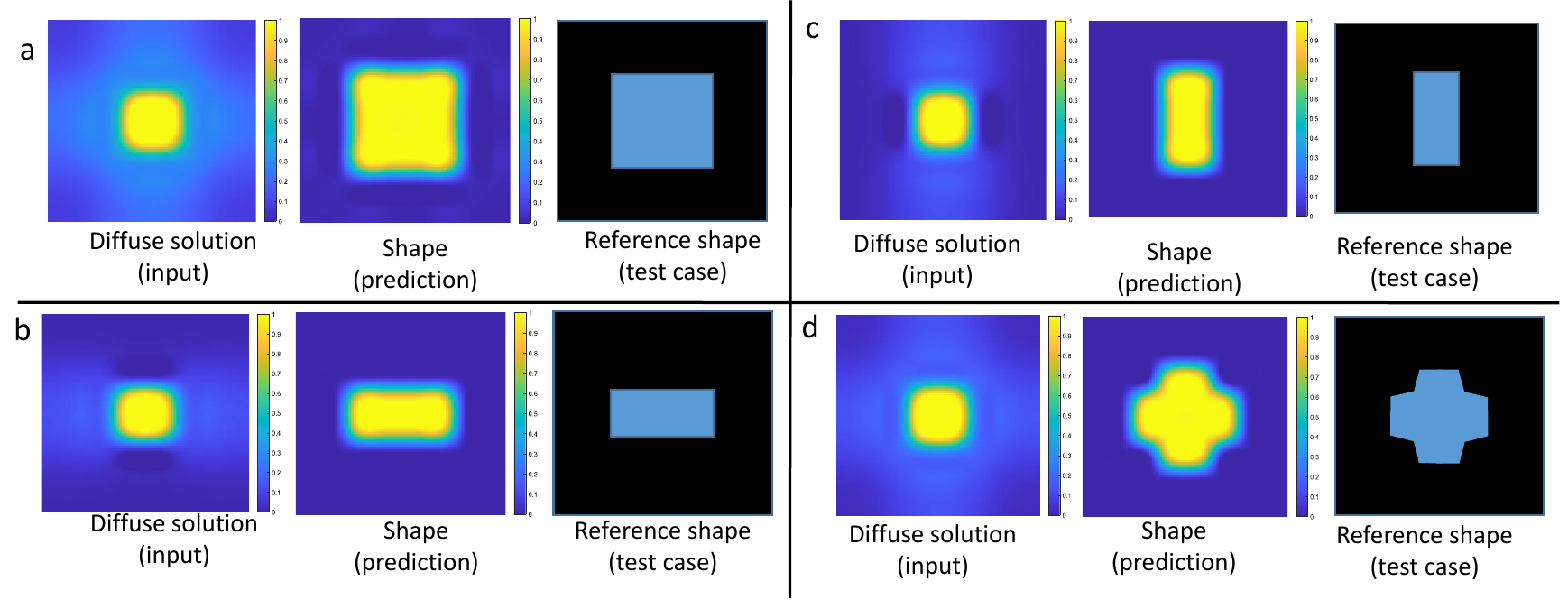}
    \caption{Reverse diffusion solutions for different initial concentration profiles. The expected precipitate shapes are indicated.}
    \label{inv}
\end{figure}

The trajectories predicted by the DLNN can be navigated both in forward and reverse directions, the latter allowing solution of inverse problems. In this example, four distinct precipitate shapes within a matrix—each with a concentration of one inside the precipitate and zero within the matrix—were utilized. Employing the same diffusion coefficients as in prior experiment, the precipitates were observed to diffuse into the matrix, with their concentration recorded at 0.02 seconds. The DLNN was initially trained on forward trajectories, after which it began its reverse operation from a diffused image at t = 0.02 seconds. By predicting the rate of change at that instant and performing time steps in the reverse direction, the DLNN was used to effectively navigated the trajectory in reverse. The resulting predictions of precipitate shapes are shown in Figure \ref{inv}. Despite the initial shapes being significantly diffused at the outset, the DLNN managed to recover the ground truth shapes by traversing the trajectory. 

\section{Discussion and Conclusions}

We have demonstrated, on the basis of Morse-Feshbach Lagrangian, a novel Lagrangian neural network that is now widely applicable to a variety of physical problems compared to the state-of-the-art which only considers mechanical equations of motion. The Morse and Feshbach approach has always been regarded as just a mathematical trick because of the arbitrary suggestion of a completely formal, non-observable space being constructed. But, this can be recognized as, what is now standard practice: lift the problem to a higher dimensional space, where the ill-posedness is removed.

Examples show that the approach can be used to extrapolate the dynamics of physical systems and is better than a baseline neural network. The method is quite general and in the future, this approach can be easily extended to other physical equations (including viscous flow, electromagnetics and Schrodinger's equation) significantly broadening our understanding of materials physics using observed data.

While we believe we have shown that the Morse--Feshbach approach works for dissipative systems, there are some limitations to bear in mind. We have developed a formalism that allowed us to not explicitly treat the mirror space dimensions, $\eta$, but there may be some advantage of constructing values for this space. In Newtonian dynamics, this space can be understood as a mirror image system with`negative friction' into which energy is drained. With a dissipative system, the state becomes progressively smoother, as more and more states are consolidated closely into a smooth state. For example, a Gaussian blur on sets of similar images, will ultimately produce an image that is representative of them all. This makes the inverse problem, necessarily ill-posed because there is no way to prefer one backward solution over another in the presence of noise. Consistent with the Morse and Feshbach original idea, these backward solutions can be differentiated by the values of $\eta$, if they were constructed. With appropriate shaping of the distribution of this space, it should be possible to use this to develop yet another generator model, only this time, it would robustly generate previous states.

\section*{Acknowledgements}
This research was supported in part by the Air Force Research Laboratory Materials and Manufacturing Directorate, through the Air Force Office of Scientific Research Summer Faculty Fellowship Program, Contract Numbers FA8750-15-3-6003 and FA9550-15-0001. This research was supported in part through computational resources and services provided by Advanced Research Computing at the University of Michigan, Ann Arbor.

\end{document}